\documentclass[runningheads]{llncs}
\usepackage[T1]{fontenc}
\usepackage{graphicx,verbatim}
\usepackage{amsmath}
\usepackage{booktabs}
\usepackage{multirow}
\usepackage[table]{xcolor}
\begin{document}
\title{Agent-Guided Relational Concept Discovery: Toward Interpretable Surgical Margin Assessment}
\titlerunning{Agent-Guided Relational Concept Discovery}

\author{
Nooshin Maghsoodi\inst{1}\ \thanks{Corresponding author.} \and
Amoon Jamzad\inst{1}\ \and
Robert Policelli\inst{1}\ \and
Mohammad Farahmand\inst{1} \and
Dilakshan Srikanthan\inst{1} \and
Martin Kaufmann\inst{1}\ \and
Kevin Y. M. Ren\inst{1}\ \and
Shaila Merchant\inst{1}\ \and
Sonal Varma\inst{1}\ \and
Ross Walker\inst{1} \and
Doug McKay\inst{1} \and
John Rudan\inst{1} \and
Gabor Fichtinger\inst{1} \and
Parvin Mousavi\inst{1}
}

%\index{Maghsoodi, Nooshin}
%\index{Jamzad, Amoon}
%\index{Policelli, Robert}
%\index{Farahmand, Mohammad}
%\index{Srikanthan, Dilakshan}
%\index{Kaufmann, Martin}
%\index{Ren, Kevin Y. M.}
%\index{Merchant, Shaila}
%\index{Varma, Sonal}
%\index{Walker, Ross}

%\index{McKay, Doug}
%\index{Rudan, John}
%\index{Fichtinger, Gabor}
%\index{Mousavi, Parvin}

\authorrunning{N. Maghsoodi et al.}

\institute{
Queen's University, Kingston, ON, Canada\\
\email{nooshin.maghsoodi@queensu.ca}
}
\maketitle

\begin{abstract}
Deep learning models can effectively use Rapid Evaporative Ionization Mass Spectrometry (REIMS) data for surgical margin assessment. However, their clinical adoption remains challenging due to limited generalization to operating room conditions. This difficulty arises because models are typically trained on labeled spectra collected from resected tissue samples, while they must operate on noisy, unlabeled data acquired directly during surgery. In addition, the black-box nature of deep learning models makes it difficult to understand and systematically improve their behavior. Concept-based learning offers a promising way to address these challenges by mapping raw measurements to human-understandable concepts. However, supervised concept-based approaches rely on concept annotations, which are difficult to obtain in complex mass spectrometry workflows. We propose Agent-Guided Concept Discovery, a framework that learns meaningful concepts directly from data without requiring predefined concept labels. During training, a reasoning agent refines semantic descriptions of the learned concepts and adaptively adjusts their weight based on diagnostic relevance. These concepts are further grounded using a biochemical knowledge graph to ensure consistency with known metabolic relationships. Across Skin and Breast Cancer datasets, our model improves balanced accuracy and sensitivity over the baseline. In a representative intraoperative case, it shows fewer false positives, indicating better generalization to surgical conditions.

% Authors must provide keywords and are not allowed to remove this Keyword section.

\end{abstract}

\section{Introduction}

Achieving clear surgical margins has a significant impact on cancer outcomes. Negative margins indicate that no malignant cells remain at the boundary of the excised tissue, suggesting that the tumor has been fully removed and reducing the risk of residual disease. Today, margin status is typically determined postoperatively through histopathology, which provides no opportunity for intraoperative adjustment. In contrast, real-time margin assessment allows surgeons to course-correct during resection, both avoiding cutting through the tumor and minimizing unnecessary removal of healthy tissue \cite{1jamzad2020improved}. Rapid Evaporative Ionization Mass Spectrometry (REIMS) enables intraoperative feedback by providing instantaneous metabolomic profiles from cauterized tissue, which can be classified using machine learning to identify cancer at the point of incision \cite{2balog2013intraoperative}. 

Prior works have explored multiple strategies to improve REIMS-based margin assessment, including Bayesian neural networks for uncertainty estimation and image-based representations of mass spectra \cite{4fooladgar2022uncertainty,5connolly2024imspect}. More recently, a foundation model pretrained on large-scale tandem mass spectrometry data, Deep Representations Empowering the Annotation of Mass Spectra (DreaMS) \cite{6bushuiev2025selfsupervised}, has been proposed and later adapted for REIMS applications \cite{7farahmand2025fact}. However, fine-tuning this foundation model on small, labeled ex vivo REIMS datasets risks overfitting and limits generalization to noisy intraoperative data. Moreover, despite strong performance, these models largely remain black boxes, offering limited insight into the biochemical drivers of their predictions.

Concept-based learning offers a promising alternative to purely end-to-end classification. Rather than mapping raw spectra directly to diagnostic labels, these models introduce an intermediate representation composed of human-understandable variables \cite{8koh2020concept}. By explicitly modeling such latent factors, predictions can be explained in terms of meaningful evidence, and decision-making becomes more robust to distribution shifts by relying on stable, high-level abstractions instead of low-level features \cite{9kim2018interpretability,10yuksekgonul2022post}. However, existing supervised frameworks depend on manually defined concept annotations, which are impractical in complex mass spectrometry workflows.

To address these challenges, we propose Agent-Guided Relational Concept Discovery, a framework for interpretable surgical margin assessment. We introduce a reasoning-agent-in-the-loop training framework that enables automatic discovery of discriminative concepts without requiring concept annotations. The agent analyzes shared spectral patterns, assigns semantic descriptions, and adaptively adjusts concept relevance. These concepts are further grounded using a knowledge graph to ensure consistency with established metabolic pathways. Our main contributions are summarized as follows:

\begin{enumerate}
    \item \textit{Agent-Guided Concept Discovery without Supervision.}  
A reasoning agent is incorporated into the training loop to help the model discover and refine meaningful latent representations from embeddings without requiring manual annotations. This is achieved through two key contributions below.
% The model identifies discriminative spectral patterns, while a reasoning agent uses these patterns, along with metabolic knowledge and classification performance, to assign semantic meaning and importance to each concept during training.

% The agent model considers the identified discriminative spectral patterns in samples with high concept activation and returns feedback about the importance of each concept accordingly. These weighted concepts are then incorporated into the loss function to guide the optimization of the training network parameters.

    \item \textit{Relational Concept Modeling via Knowledge Graph Construction}  
The agent receives the most informative spectral patterns associated with each learned concept and then queries metabolic databases to ground these concepts in biological knowledge. These links generate a knowledge graph connecting concepts to metabolites, pathways, and tissue states, which evolves during training to capture relationships and support more consistent agent feedback.

    \item \textit{Reasoning-Driven Concept Weight Optimization.} During training, the agent aggregates supporting evidence to provide feedback on the relevance of the learned concepts. This feedback is incorporated into an auxiliary alignment loss, which guides the learning of the concept weights. 

\end{enumerate}

\section{Materials and Methods}
\begin{figure}[!t]
\centering
\includegraphics[width=\textwidth]{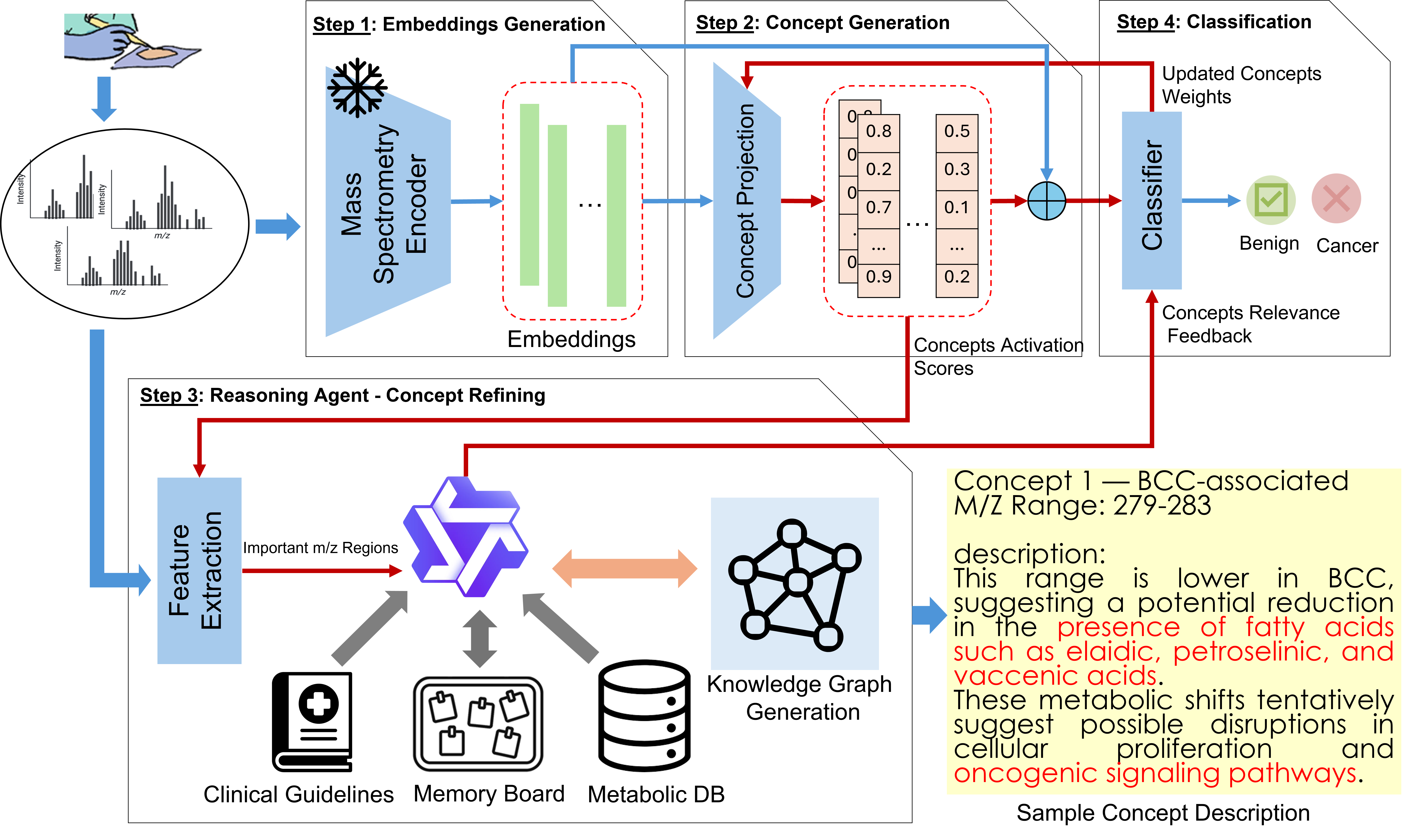}
\caption{REIMS spectra are encoded to obtain latent embeddings, which are mapped to concept activations. Samples with high concept activation are analyzed to identify discriminative spectral features. A reasoning agent integrates this information with biochemical knowledgebases to ground the learned concepts, construct a relational knowledge graph, and iteratively refine concepts during training.} 
\label{fig1}
\end{figure}

\subsection{Data}

\paragraph{Ex vivo data:}
Ex vivo REIMS spectra were collected from two surgical oncology cohorts under institutional ethics approval. The basal cell carcinoma (BCC) dataset includes 693 annotated burns from 91 patients (252 tumors, 441 benign), acquired from freshly excised specimens. The breast cancer dataset comprises 144 ex vivo burns from 11 patients (41 tumors, 103 benign), sampled from malignant tissue and adjacent benign parenchyma using the same REIMS protocol.

\paragraph{Intraoperative data:}
Intraoperative reims data were collected during a breast-conserving surgery. Over a 27-minute procedure, 1,616 spectra were recorded at 1~\textit{Hz} and labeled using intraoperative surgical annotations.

\subsection{Proposed Model}
Fig.~1 provides an overview of the proposed framework and its main components. The model is designed to jointly perform surgical margin classification while learning interpretable concepts based on a biochemical knowledge base. The overall pipeline consists of the following components.

\subsubsection{Step 1: Embedding Generation via Foundation Model}
In the first step, we use the DreaMS foundation model~\cite{6bushuiev2025selfsupervised} to process raw REIMS spectra.
% DreaMS is a large transformer-based model pre-trained on millions of tandem mass spectrometry spectra, enabling it to capture chemical patterns across different experimental datasets. 
Due to the limited size of our labeled REIMS dataset, the DreaMS parameters are kept frozen and used solely as a feature extractor. For each input spectrum $x_i$, the model produces a $d$-dimensional embedding, $z_i$, which serves as the input to the concept embedding layer.
\begin{equation}
z_i = f_{\text{DreaMS}}(x_i) \in {R}^d.
\end{equation}

\subsubsection{Step 2: Concept Embedding Layer}
The embedding $z_i$ is mapped to a set of $K$ latent concepts through a learnable linear projection layer, forming an explicit concept bottleneck. Each concept $k$ is parameterized by a weight vector $w_k \in {R}^d$ and a bias term $b_k$. For a given sample $i$, the concept activation score is computed and mapped to a bounded activation value as
\begin{equation}
s_{ik} = w_k^\top z_i + b_k, \qquad
p_{ik} = \sigma(s_{ik}),
\label{eq:concept_activation}
\end{equation}
where $p_{ik}$ represents the relative presence of concept $k$ in the input spectrum.

To encourage separation between concept directions, we apply a regularization term based on cosine similarity between ${w_k}_{k=1}^K$. This term discourages highly similar concept vectors, reducing redundancy while allowing related concepts to remain correlated when supported by the data.

\subsubsection{Step 3: Reasoning Agent for Concept Refinement}
To convert concept activations into biologically meaningful representations, we integrate a reasoning agent into the training loop.

\textit{Discriminative spectral region identification.}
 For each concept $k$, the feature extraction module contrasts spectra with high and low concept activations to identify \emph{m/z} regions that show discriminative intensity differences. Through this process, concepts that were previously defined only as numerical parameters in Step~2 become associated with specific \emph{m/z} ranges.

\textit{Biochemical grounding.}
The agent analyzes the identified \emph{m/z} regions by querying biochemical knowledge bases to retrieve candidate metabolites and pathways. We use RaMP~\cite{11braisted2023rampdb,12zhang2018ramp} as the primary resource, which integrates metabolite identifiers, pathways, and chemical annotations across curated databases. The agent is additionally guided by text-based domain guidelines distilled from biochemical and clinical free resources and some literature~\cite{fu2020lipid,thomsen2026msml,13kaufmann2024testing,14brorsen2024bcc}.

\textit{Memory and knowledge graph construction.}
During training, the agent maintains a memory of concept descriptions, extracted spectral ranges, relevance feedback, and classification performance across epochs. Using this information, it updates a concept-level knowledge graph linking concepts to metabolites, pathways, and tissue pathological states, which is referenced to assess relationships and refine the concept descriptions.

\textit{Description and relevance feedback.}
Based on this analysis, the agent provides (i) \textit{textual descriptions} that associate concepts with putative metabolic classes, and (ii) \textit{concept relevance} as directional feedback (\emph{more relevant}, \emph{less relevant}) based on biochemical consistency, historical trends from the memory board, and impact on classification. 

\subsubsection{Step 4: Agent-Guided Concept Alignment Loss}

As mentioned, the model discovers $K$ meaningful concepts from the training data. 
However, in biochemical analysis, different spectral patterns or \emph{m/z} regions contribute unequally to tissue characterization and diagnosis~\cite{13kaufmann2024testing,14brorsen2024bcc}. 
Motivated by this, we allow each learned concept to have a distinct weight on the final prediction.

To explicitly control the contribution of each concept, we associate it with a learnable scalar weight parameter $\alpha_k$. 
To ensure interpretability and numerical stability, concept weight parameters are constrained to the range $[0,1]$ using a sigmoid reparameterization. 
The gated concept signal, $\tilde{s}_{ik}$, is then defined as
\begin{equation}
\alpha_k = \sigma(a_k), \quad a_k \in {R}, 
\qquad
\tilde{s}_{ik} = \alpha_k \, s_{ik}.
\end{equation}

The classifier receives both the global embedding and the gated concept signals as input. 
Specifically, the predicted output for sample $i$ is given by
\begin{equation}
\hat{y}_i = g\!\left( \left[\, z_i \;\; \tilde{s}_{i1} \;\; \tilde{s}_{i2} \;\; \dots \;\; \tilde{s}_{iK} \,\right] \right),
\end{equation}
where $g(\cdot)$ represents the classification head.
To align the model’s use of concepts with the reasoning agent’s feedback, we introduce an alignment loss based on a perturbation-based relevance measure inspired by~\cite{15pang2024integrating}. For each concept $k$, we measure the change in prediction when its contribution is removed from the classifier input. This value reflects how strongly the model’s decision depends on concept $k$.

% To encourage the model to rely on concepts consistent with feedback from the reasoning agent, we introduce an alignment term in the loss function. 
% Inspired by prior work on clinically guided concept-based models~\cite{15pang2024integrating}, we quantify the model’s reliance on each concept using a perturbation-based relevance measure. 
% For each concept $k$, we compute the change in prediction when its contribution is removed from the classifier input:
\begin{equation}
\Delta Y_k = \left| \hat{y}_i - \hat{y}_{i\,(\tilde{s}_{ik} \rightarrow 0)} \right|.
\end{equation}

Based on the reasoning agent’s feedback, concepts are grouped into high-relevant concepts $\mathcal{K}_H$ and low-relevant concepts $\mathcal{K}_L$. 
The alignment loss encourages increased reliance on concepts deemed clinically meaningful by the agent while decreasing dependence on less informative concepts:
\begin{equation}
\mathcal{L}_{\text{align}} =
\sum_{k \in \mathcal{K}_H} (1 - \Delta Y_k)
+
\sum_{k \in \mathcal{K}_L} \Delta Y_k .
\end{equation}

The final objective combines classification loss with agent-guided alignment, where $\mathcal{L}_{\text{class}}$ is binary cross-entropy and $\lambda \geq 0$ controls the alignment strength.
\begin{equation}
\mathcal{L}_{\text{total}} =
\mathcal{L}_{\text{class}}
+
\lambda \, \mathcal{L}_{\text{align}}.
\end{equation}

\begin{table}[t]
\caption{Classification performance (mean $\pm$ standard deviation over 30 runs) on REIMS datasets.}
\label{tab:classification_results}
\centering
{\fontsize{8}{8}\selectfont
\renewcommand{\arraystretch}{1.15}
\setlength{\tabcolsep}{4pt}
\begin{tabular}{l l c c c c}
\hline
\textbf{Data} & \textbf{Method} & \textbf{Bal. Acc.} & \textbf{Sens.} & \textbf{Spec.} & \textbf{AUROC} \\
\hline
\multirow{3}{*}{\textbf{Skin Cancer}} 
& Transformer 
& $0.74\!\pm\!0.027$ 
& $\mathbf{0.72}\!\pm\!0.111$ 
& $0.76\!\pm\!0.084$ 
& $0.85\!\pm\!0.015$ \\
& DreaMS~\cite{6bushuiev2025selfsupervised}
& $0.71\!\pm\!0.034$ 
& $0.59\!\pm\!0.086$ 
& $\mathbf{0.83}\!\pm\!0.054$ 
& $0.83\!\pm\!0.014$ \\
& \textbf{Our method} 
& $\mathbf{0.76\!\pm\!0.026}$ 
& $0.70\!\pm\!\mathbf{0.043}$ 
& $0.82\!\pm\!\mathbf{0.031}$ 
& $\mathbf{0.86\!\pm\!0.013}$ \\
\hline
\multirow{3}{*}{\textbf{Breast Cancer}} 
& Transformer 
& $0.82\!\pm\!0.035$ 
& $0.78\!\pm\!0.216$ 
& $0.86\!\pm\!0.062$ 
& $0.93\!\pm\!0.021$ \\
& DreaMS~\cite{6bushuiev2025selfsupervised}
& $0.84\!\pm\!0.028$ 
& $0.80\!\pm\!0.051$ 
& $0.89\!\pm\!0.026$ 
& $0.96\!\pm\!0.012$ \\
& \textbf{Our method} 
& $\mathbf{0.87\!\pm\!0.018}$ 
& $\mathbf{0.81\!\pm\!0.044}$ 
& $\mathbf{0.93\!\pm\!0.024}$ 
& $\mathbf{0.98\!\pm\!0.013}$ \\
\hline
\end{tabular}}
\end{table}

\subsection{Experiments}

Our experiments consist of the following evaluations: (i) \textit{Classification performance comparison.} 
We compare the proposed framework against a DreaMS-based baseline that performs classification on foundation model embeddings, on two REIMS mass spectrometry datasets. (ii) \textit{Concept quality evaluation.} 
We assess interpretability and biological relevance by examining how concept activations vary with hormone receptor status and align with established metabolic pathways. (iii) \textit{Generalization to intraoperative data.} We assess by evaluating performance on an in vivo REIMS case acquired during surgery. (iv) \textit{Ablation studies.} 
We analyze the contribution of key components by ablating biochemical knowledge base integration and knowledge graph grounding. 

\paragraph{Implementation Details.}
The reasoning agent uses the \texttt{Qwen2.5-7B-Instruct} model. The model learns $K=8$ concepts and is trained using Adam with a learning rate of $10^{-3}$ and a batch size of $32$ for up to 40 epochs. Datasets are split into training, validation, and test sets, hyperparameters are selected on the validation set, and all experiments are repeated 30 times to report mean and standard deviation. Full code and configurations will be publicly available after acceptance.

\section{Results and Discussion}
\subsubsection{Classification performance comparison}
The classification results are shown in Table~\ref{tab:classification_results}. Overall, the proposed model outperforms the DreaMS baseline across both datasets. This indicates that incorporating agent-guided concept embeddings, with the goal of improving generalization and interpretability, does not sacrifice classification performance and can even improve it. On the Skin Cancer dataset, we observe an approximate 7\% relative improvement in balanced accuracy, along with higher AUROC and a noticeable increase in sensitivity, suggesting improved detection of tumor tissue.
Similarly, on the Breast Cancer dataset, our method achieves higher balanced accuracy (0.87) while also improving both sensitivity and specificity. 

\subsubsection{Clinical Evaluation of Concepts}

\begin{figure}[!t]
\centering
\includegraphics[width=\textwidth]{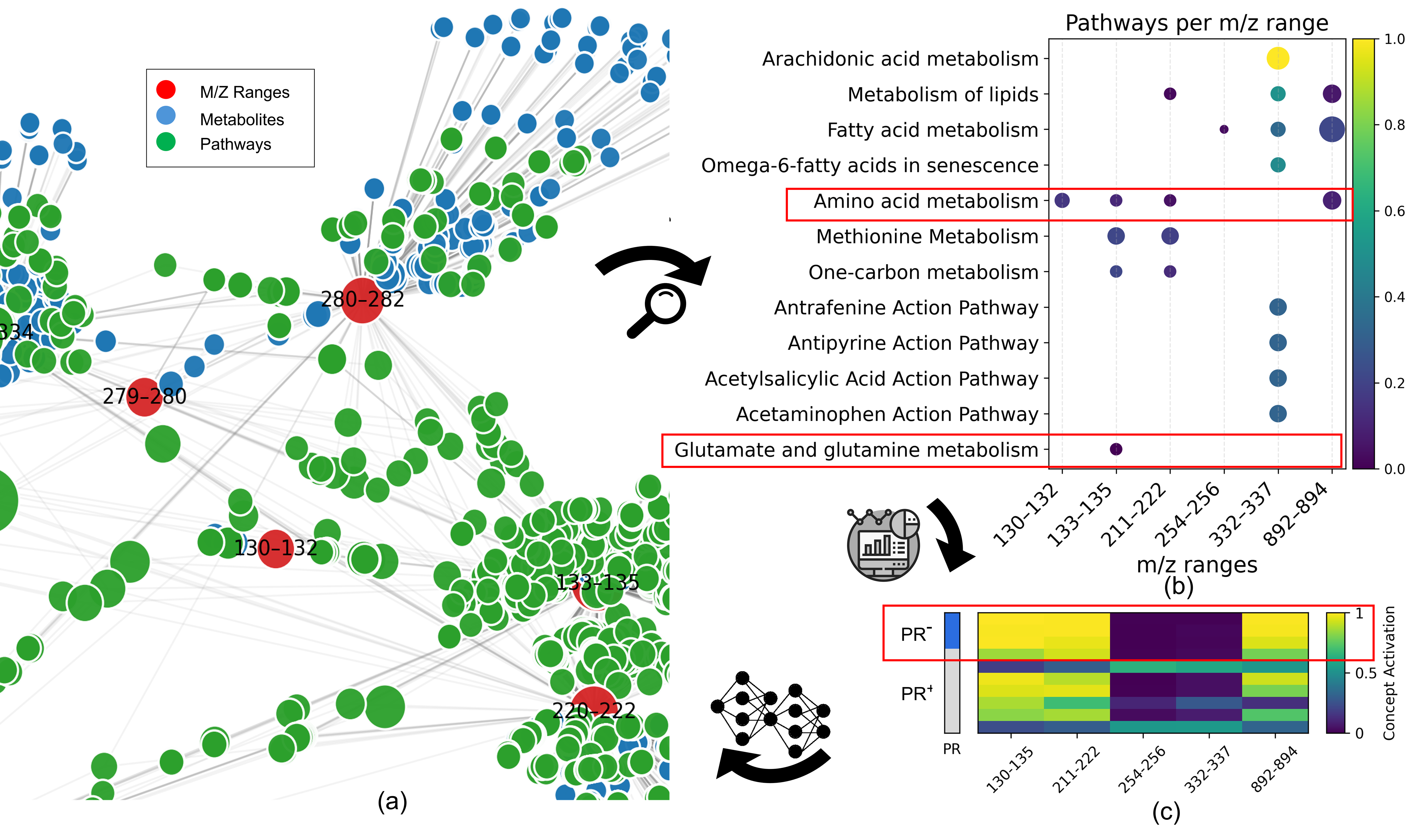}
\caption{(a) Breast cancer knowledge graph linking m/z ranges, metabolites, and pathways.
(b) Zoomed-in selected pathways for some m/z ranges in (a).
(c) Concept activations across tumor samples with their Progesterone Receptor (PR) status.} 
\label{fig4}
\end{figure}

Fig.~\ref{fig4} illustrates how the learned concepts are grounded in biochemical knowledge. Fig.~\ref{fig4}(a) shows a portion of the Breast Cancer knowledge graph linking informative \emph{m/z} ranges to candidate metabolites and pathways discovered during training. A zoomed-in view from this knowledge graph is shown in Fig.~\ref{fig4}(b), highlighting selected \emph{m/z} ranges and some of their associated pathways, enabling assessment of clinical relevance.
We evaluated this relevance by analyzing concept activation scores for tumor samples in the Breast Cancer test set. Fig.~\ref{fig4}(c) shows the activation of selected concepts together with progesterone receptor (PR) status. Prior studies~\cite{16shahnazari2025multimodal,17budczies2015glutamate} report that PR-negative breast cancers are associated with glutamine metabolism and increased amino acids. Consistent with these findings, Fig.~\ref{fig4}(b) highlights amino acid– and glutamine-related pathways linked to the \emph{m/z} subbands 130–135, 211–222, and 892–894, while Fig.~\ref{fig4}(c) shows higher activation of these subbands in PR-negative samples. This consistency supports the biological relevance of the learned concepts and demonstrates the interpretability enabled by the generated knowledge graph.

% Fig.~\ref{fig4} illustrates how the learned concepts are grounded in biochemical knowledge. Fig.~\ref{fig4}(a) shows a portion of the BCC knowledge graph, where informative \emph{m/z} ranges are linked to candidate metabolites and metabolic pathways identified during training. A zoomed-in example from a similar knowledge graph for Breast Cancer is shown in Fig.~\ref{fig4}(b), highlighting some selected \emph{m/z} ranges and some of their associated pathways. This view enables further evaluation of the clinical relevance of the learned concepts. To assess this relevance, we analyzed concept activation scores for tumor samples in the Breast Cancer test set. Fig.~\ref{fig4}(c) shows the activation of selected concepts alongside the progesterone receptor (PR) status of each sample. Prior studies~\cite{16shahnazari2025multimodal,17budczies2015glutamate} report that PR-negative breast cancers exhibit increased amino acid metabolism compared to PR-positive tumors. Consistent with this, Fig.~\ref{fig4}(b) highlights amino acid–related pathways associated with the \emph{m/z} subbands 130–135, 211–222, and 892–894. Fig.~\ref{fig4}(c) further shows that these same subbands have higher activation in PR-negative samples. This agreement with established clinical findings supports the biological meaning of the learned concepts and demonstrates the interpretability of the proposed framework using the generated knowledge graph.

\begin{figure}[!t]
\centering
\includegraphics[width=\textwidth]{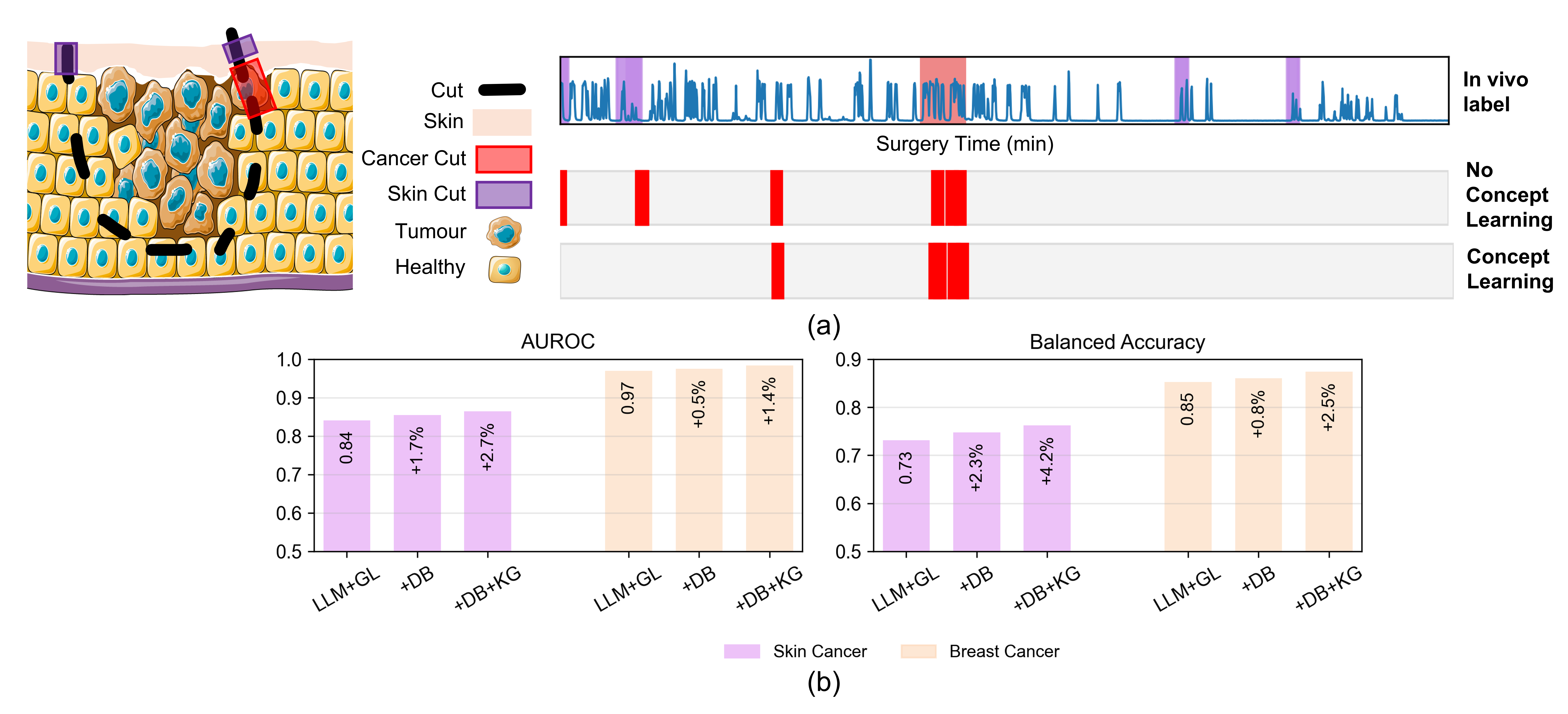}
\caption{(a) Intraoperative breast cancer example. The x-axis denotes surgery time. Red regions indicate surgeon-noted tumors, while yellow regions correspond to skin cuts. (b) Ablation study on both datasets, as metabolic database querying (DB) and knowledge graph reasoning (KG) are added to support the agent + clinical guidelines (GL). Values indicate relative performance gains.} 
\label{fig2}
\end{figure}

\subsubsection{Generalization to intraoperative data}
% Fig.~\ref{fig2} illustrates the model’s behavior on a representative intraoperative breast cancer sample, in which ground-truth labels are provided via surgeon call-outs during surgery. In this example, the red region in the first row corresponds to tissue identified by the surgeon as tumor, while the yellow regions indicate skin, which exhibits some molecular signatures similar to tumor when cut with the reims. Model predictions depend on the decision threshold applied to the classifier output probability. In practice, this threshold can be adjusted based on surgical context, for example, using a lower threshold when operating close to the tumor and a higher threshold in regions farther away. In this experiment, where both models achieve a sensitivity of 1.0 for tumor detection, we observe that the model without concept integration produces a higher number of false positives in comparison to the concept-enhanced model. This suggests that incorporating biologically aware concepts helps the model better distinguish tumor tissue from skin.

Fig.~\ref{fig2}(a) shows model predictions on a representative intraoperative breast cancer case with surgeon-provided call-out labels. Red overlayed regions denote the tumor, and yellow regions denote the skin during incision. While both models achieve perfect tumor sensitivity, the concept-enhanced model produces fewer false positives than the baseline, particularly in skin regions with tumor-like signatures. This suggests that incorporating biologically aware concepts helps the model better distinguish tumor tissue from skin.

\subsubsection{Ablation Study}
Ablation study results are shown in Fig.~\ref{fig2}(b). In this experiment, components that support the reasoning agent are added incrementally to analyze the effect of the metabolic database and the knowledge graph. Incorporating the metabolic database improves classification performance, with additional gains observed when the knowledge graph is introduced. These effects are more observable on the skin dataset, where the largest relative improvements are observed in balanced accuracy, indicating improved sensitivity to tumor-related spectral patterns. We believe that adding the guideline component alone primarily helps reduce hallucinated semantic explanations generated by the agent.

% \begin{figure}[!t]
% \centering
% \includegraphics[width=\textwidth]{fig3.png}
% \caption{Ablation study. Metrics are reported on both datasets as agent supporting modules are progressively added. clinical guidelines (GL), metabolic database querying (DB), and knowledge graph reasoning (KG). Values indicate relative performance gains.} 
% \label{fig3}
% \end{figure}

\section{Conclusion}

In our proposed framework, an agent operates within the training loop to help the model discover meaningful concepts without requiring explicit concept annotations, with the goals of improved interpretability and stronger generalization. This approach extends models trained on ex vivo data to challenging intraoperative settings while making their decisions easier to understand. The model learns high-level metabolic concepts and directly integrates them into the prediction process, grounding decisions in patterns that remain stable under intraoperative noise. In addition, a dynamically constructed, data-specific knowledge graph reveals underlying biochemical structure and supports iterative interpretation and concept refinement beyond what fixed rule-based systems can provide. A current limitation is the reliance on general metabolic database queries; future work will explore more targeted querying strategies and the integration of gene-level information to further enhance biological specificity and interpretability.

\begin{credits}
\subsubsection{\ackname} This work was supported in part by the Natural Sciences and Engineering Research Council of Canada (NSERC); the Canadian Institutes of Health Research (CIHR); and the Vector Institute. The work of Parvin Mousavi was supported in part by a Canada CIFAR AI Chair and a Canada Research Chair.

\subsubsection{\discintname}
The authors have no competing interests.
\end{credits}

% Our agent-guided framework extends models trained on ex vivo data to challenging intraoperative settings, while also making model decisions easier to understand. The model learns and refines high-level metabolic concepts that capture the most informative spectral regions for classification and directly incorporates these concepts into the prediction process. By grounding decisions in biologically meaningful concepts that remain stable under intraoperative noise, the model improves generalization without sacrificing accuracy. In addition, the dynamically constructed, data-specific knowledge graph reveals useful biochemical structure in the data and supports both interpretation and concept refinement. The reasoning agent continuously integrates new evidence and goes beyond what fixed rule-based systems can achieve. Overall, our approach preserves classification performance while providing biologically meaningful explanations, with improved generalization demonstrated in an intraoperative case.

% %
% % ---- Bibliography ----
% %
% % BibTeX users should specify bibliography style 'splncs04'.
% % References will then be sorted and formatted in the correct style.
% %
\bibliographystyle{splncs04}
\bibliography{references}




% LNCS Homepage, \url{http://www.springer.com/lncs}, last accessed 2023/10/25
% \end{thebibliography}
\end{document}